\documentclass[11pt]{article}
\usepackage{times,epsfig,boxedminipage,url,alltt}
\usepackage{natbib}
\bibpunct[:]{(}{)}{;}{a}{}{,}

\usepackage{fancyheadings}
\def\UrlFont{\small\tt}

\def\elt#1{{\small\sf #1}}
\def\attr#1{{\small\sf #1}}
\def\code#1{{\small\sf #1}}

\title{Extending Dublin Core Metadata to\\
Support the Description and Discovery\\
of Language Resources}

\author{Steven Bird$^{\ast}$ and Gary Simons$^{\dagger}$\\
$^{\ast}$University of Melbourne and University of Pennsylvania\\
$^{\dagger}$SIL International\\
{\small Email: \texttt{sb@ldc.upenn.edu}, \texttt{Gary\_Simons@sil.org}}
}

\date{2003}

\begin{document}
\maketitle

\begin{abstract}
As language data and associated technologies proliferate and
as the language resources community expands,
it is becoming increasingly difficult to locate and reuse existing
resources.  Are there any lexical resources for such-and-such a language?
What tool works with transcripts in this particular format?
What is a good format to use for linguistic data of this type?
Questions like these dominate many mailing lists, since web search engines are
an unreliable way to find language resources.
This paper reports on a new digital infrastructure for discovering
language resources being developed by the Open Language Archives Community
(OLAC).  At the core of OLAC is its metadata format, which is designed
to facilitate description and discovery of all kinds of language resources,
including data, tools, or advice.  The paper describes OLAC metadata,
its relationship to Dublin Core metadata, and its dissemination using
the metadata harvesting protocol of the Open Archives Initiative.
\end{abstract}

\section{Introduction}

Language technology and the linguistic sciences are
confronted with a vast array of \emph{language resources},
richly structured, large and diverse.
Multiple \emph{communities} depend on language resources, including
linguists, engineers, teachers and actual speakers.
Many individuals and institutions provide key pieces of the infrastructure,
including archivists, software developers, and publishers.
Today we have unprecedented opportunities to \emph{connect}
these communities to the language resources they need.
First, inexpensive mass storage technology permits large resources to
be stored in digital form, while
the Extensible Markup Language (XML) and Unicode provide flexible
ways to represent structured data and ensure its long-term survival.
Second, digital publication -- both on and off the world wide web --
is the most practical and efficient means of sharing language resources.
Finally, a standard resource description model, the Dublin Core Metadata
Set, together with an interchange method provided by the Open Archives
Initiative (OAI), make it possible to construct a union catalog over multiple
repositories and archives.

In December 2000, a new initiative which applied the OAI to language
archives was founded, with the following statement of purpose:

\begin{quote}
OLAC, the Open Language Archives Community, is an international partnership
of institutions and individuals who are creating a worldwide virtual
library of language resources by: (i)~developing consensus on best current
practice for the digital archiving of language resources, and
(ii)~developing a network of interoperating repositories and services for
housing and accessing such resources.
\end{quote}

This paper presents the motivation and governing ideas of OLAC,
Dublin Core metadata and the Open Archives Initiative Protocol
for Metadata Harvesting (\S\ref{sec:background}),
followed by the OLAC Metadata Set (\S\ref{sec:metadata}).
It concludes with an overview of ongoing developments and
a call for participation by the wider community.
Updated information on OLAC is available from the OLAC Gateway
[\url{www.language-archives.org}].

\section{Locating Data, Tools and Advice}
\label{sec:background}

We can observe that the
individuals who use and create language resources
are looking for three things: data, tools, and advice.
By DATA we mean any information that documents or describes a language,
such as a published monograph, a computer data file, or
even a shoebox full of hand-written index cards. The information could range
in content from unanalyzed sound recordings to fully transcribed and annotated
texts to a complete descriptive grammar. 
By TOOLS we mean computational resources that facilitate creating, viewing,
querying, or otherwise using language data. Tools include not just software
programs, but also the digital resources that the programs depend on, such as
fonts, stylesheets, and document type definitions.
By ADVICE we mean any information about
what data sources are reliable, what tools are appropriate in a given
situation, what practices to follow when creating new data, and so forth
\citep{BirdSimons03language}.
In the context of OLAC, the term \emph{language resource} is broadly
construed to include all three of these: data, tools and advice.

\begin{figure}[b]
\centerline{\includegraphics[width=0.9\linewidth]{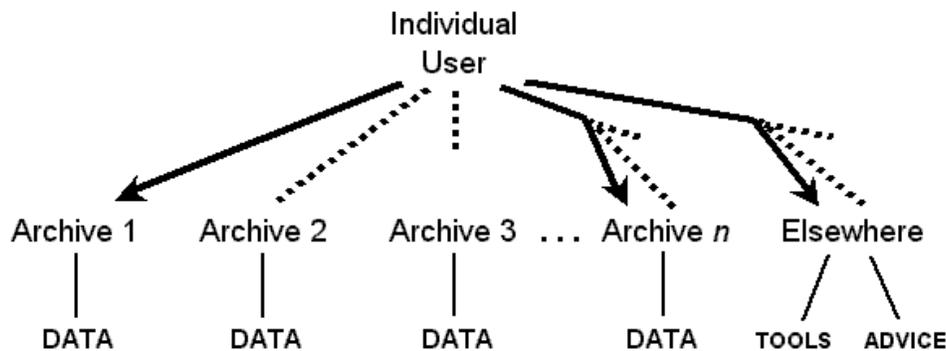}}
\caption{In reality the user can't always get there from here}
\label{fig:vision2}
\end{figure}
Unfortunately, today's user does not have ready access to the resources
that are needed. Figure~\ref{fig:vision2}
offers a diagrammatic view of the reality.
Some archives (e.g. Archive 1) do have a site on the internet which the user is
able to find, so the resources of that archive are accessible. Other archives
(e.g. Archive 2) are on the internet, so the user could access them in theory,
but the user has no idea they exist so they are not accessible in practice.
Still other archives (e.g. Archive 3) are not even on the internet. And there
are potentially hundreds of archives (e.g. Archive $n$) that the user
needs to know about. Tools and advice are out there as well, but are at many
different sites.

There are many other problems inherent
in the current situation. For instance, the user may not be able to find all
the existing data about a language of interest because different sites have
called it by different names (low recall).
The user may be swamped with irrelevant resources because search terms
have important meanings in other domains (low precision).
(For a detailed discussion of precision and recall in the context of
metadata, see \cite{Svenonius00}.)
The user may not be able to use an accessible
data file for lack of being able to match it with the right tools. The user may
locate advice that seems relevant but have no basis for judging its merits.

\subsection{Bridging the gap}

\subsubsection{Why improved web-indexing is not enough}

As the internet grows and web-indexing technologies improve one might hope
that a general-purpose search engine should be sufficient to bridge the gap
between people and the resources they need.  However, this is a vain hope.
The first reason is that many language resources, such as audio files
and software, are not text-based.  The second
reason concerns language identification, the single most important
property for describing language resources.  If a language has a canonical name
which is distinctive as a character string, then the user has a chance of
finding any online resources with a search engine.
However, the language may have
multiple names, possibly due to the vagaries of romanization, such as a
language known variously as Fadicca, Fadicha, Fedija, Fadija, Fiadidja,
Fiyadikkya, and Fedicca (giving low recall).
The language name may collide with a word which has
other interpretations that are vastly more frequent, e.g.\ the language
names Mango and Santa Cruz (giving low precision).

The third reason why general-purpose search engines are inadequate is
the simple fact that much of the material is not,
and will not, be documented in free prose on the web.
Either people will build systematic catalogues of their resources,
or they won't do it at all.
Of course, one can always export a back-end database
as HTML and let the search engines index the materials.
Indeed, encouraging people to document resources and make them
accessible to search engines is part of our vision.
However, despite the power of web search engines, there remain many
instances where people still prefer to use more formal databases to
house their data.

This last point bears further consideration.  The challenge is to
build a system for ``bringing like things together and differentiating among
them'' \citep{Svenonius00}.
There are two dominant storage
and indexing paradigms, one exemplified by traditional databases and one
exemplified by the web.  In the case of language resources, the metadata is
coherent enough to be stored in a formal database, but sufficiently
distributed and dynamic that it is impractical to maintain it centrally.
Language resources occupy the middle ground between the two paradigms, neither of which
will serve adequately.  A new framework is required that permits the best of
both worlds, namely bottom-up, distributed initiatives, along with consistent,
centralized finding aids.  The Dublin Core Metadata Initiative and the
Open Archives Initiative provide the framework we need to ``bridge the gap.''

\subsubsection{The Dublin Core Metadata Initiative}

The Dublin Core Metadata Initiative began in 1995 to develop
conventions for resource discovery on the web [\url{dublincore.org}].
The Dublin Core (DC) metadata elements represent a broad,
interdisciplinary consensus about the core set of elements that are
likely to be widely useful to support resource discovery.  The Dublin
Core consists of 15 metadata elements, where each element is optional
and repeatable: \elt{title, creator, subject, description, publisher,
  contributor, date, type, format, identifier, source, language,
  relation, coverage, rights}.  This set can be used to describe
resources that exist in both digital and traditional formats.

To support more precise description and more focussed searching, the DC
metadata set has been extended with encoding schemes and refinements
\citep{DCQ00,DCER02}.
An encoding scheme specifies a particular controlled vocabulary or notation
for expressing the value of an element.  An encoding scheme serves to aid a
client system in interpreting the exact meaning of the element content. A
refinement makes the meaning of the element more specific.
For example,
a \elt{language} element can be {\it encoded}
using the conventions of RFC 3066 to unambiguously identify the language
in which the resource is written (or spoken).
A \elt{subject} element can be given a language {\it refinement}
to restrict its interpretation to concern the language the resource is about.

\subsubsection{The Open Archives Initiative}

The Open Archives Initiative (OAI)
was launched in October 1999 to provide a common framework across
electronic preprint archives, and it has since been broadened
to include digital repositories of scholarly materials regardless
of their type
[\url{www.openarchives.org}] \citep{LagozeVandeSompel01,VandeSompelLagoze02}.
Each participating archive, or ``data provider,''
has a network accessible server offering public access
to metadata records describing archive holdings.  The holdings
themselves may be documents, raw data, software,
recordings, physical artifacts, digital surrogates, and so forth.
Each metadata record describes an archive holding, and includes
a reference to an entry point for the holding such as a URL or a
physical location.

Participating archives must comply
with two standards: the {\it OAI Shared Metadata Set} (Dublin Core) which
facilitates interoperability across all repositories participating in the
OAI, and the {\it OAI Protocol for Metadata Harvesting} which allows
``service providers'' to combine metadata from multiple archives into
a single catalogue.  End-users interact directly with a service provider
to quickly locate distributed resources.

\subsection{Applying the OAI to language resources using specialized metadata}

The OAI infrastructure is a new invention:
it has the bottom-up, distributed character of the web,
while simultaneously having the efficient, structured
nature of a centralized database.  This combination is well-suited to
the language resource community, where the available data is growing
rapidly and where a large user-base is fairly consistent in how it describes
its resource needs.

Recall that the OAI community is defined by the archives which
comply with the OAI metadata harvesting protocol
and that register with the OAI.
Any compliant repository can register as an OAI archive, and
the metadata provided by the archive is open to the public.
OAI data providers may support metadata formats in addition to DC.
A specialist community can define a metadata format
specific to its domain and expose it via the OAI protocol.
Service providers, data providers and users that
employ this specialized metadata format constitute an OAI \emph{subcommunity}.

Consequently, applying the OAI to language resources is chiefly a matter of
having a common metadata format tailored for
language resource description and discovery.
Section~\ref{sec:metadata} reports on such a format, which is
already in use by over twenty archives
having a combined total of 30,000 metadata records.
These OLAC metadata records can be harvested from multiple archives
using the OAI protocol and stored in a single location, where end-users can query
all participating archives simultaneously.
The LINGUIST List now offers an OLAC cross-archive search
service at [\url{http://www.linguistlist.org/olac}].

\section{A Core Metadata Set for Language Resources}
\label{sec:metadata}

The OLAC Metadata Set extends the Dublin Core set only to
the minimum degree required to express basic properties
of language resources which are useful as finding aids.
All Dublin Core elements and refinements are used in the OLAC Metadata Set. In
order to meet the specific needs of the language resources community, certain
elements have been extended following DCMI guidelines
\citep{DCQ00,DCXML03}.
This section describes some of
the attributes, elements and controlled vocabularies of
the OLAC Metadata Set, then shows how they are represented in XML
and how they are mapped to other formats for wider dissemination.

\subsection{Attributes used in implementing the OLAC Metadata Set}

Three attributes -- \attr{type}, \attr{code}, and \attr{lang}
are used throughout the XML implementation
of the metadata elements.
The \attr{type} attribute is used to qualify the Dublin Core element,
by refining its meaning (to make it narrower or more specific),
or by identifying an encoding scheme, or both.  If the \attr{type} specifies
one of the OLAC vocabularies, then the \attr{code} attribute is used
to hold the selected value.  For example, with the \elt{subject} element,
we may specify the type \attr{olac:language} to indicate that we are
describing the subject language of the resource.  We may also provide
a code \attr{x-sil-BAN} to uniquely identify the language.  We may further
supply element content, as a freeform elaboration of the coded value.
This design permits service providers to uniformly interpret the
meaning of any code value, thereby providing good precision and recall.
At the same time, data providers may use the element content when there is
not an appropriate code or when they want to add qualifications to the
coded value.

As with Dublin Core, every element in the OLAC metadata set may use
the \attr{lang} attribute. It specifies the language in which the text
in the content of the element is written.  By using multiple instances
of the metadata elements tagged for different languages, data
providers may offer their metadata records in multiple languages.

\subsection{The elements of the OLAC Metadata Set}

In this section we present a synopsis of the elements of the OLAC metadata
set.  For each element, we provide a one sentence definition followed by a
brief discussion, systematically borrowing and adapting the definitions
provided by the Dublin Core Metadata Initiative \citep{DCER02}.  Each element
is optional and repeatable.

\begin{description}\setlength{\itemsep}{0pt}\setlength{\parskip}{0pt}

\item[\elt{contributor}:]
{\bf An entity responsible for making contributions to the content
of the resource.}
Examples of a Contributor include a person, an organization, or a
service.  Recommended best practice is to identify the role
played by the named entity in
the creation of the resource using the OLAC Role Vocabulary
\citep{OLAC-Role}.
      
\item[\elt{coverage}:]
{\bf The extent or scope of the content of the resource.}
Coverage will typically include spatial location or temporal period.
Where the geographical information is predictable from the language identification,
it is not necessary to specify geographic coverage.

\item[\elt{creator}:]
{\bf An entity primarily responsible for making the content of the resource.}
As with the \elt{contributor} element,
recommended best practice is to identify the role
played by the named entity in
the creation of the resource using the OLAC Role Vocabulary
\citep{OLAC-Role}.

\item[\elt{date}:]
{\bf A date associated with an event in the life cycle of the resource.}
Best practice is to use the W3C Date and Time Format \citep{W3CDTF}.
Dublin Core qualifiers may be used to refine the meaning
of the date (for instance, date of
creation versus date of issue versus date of modification, and so on). The
refinements to \elt{date} are defined in \citep{DCER02}.

\item[\elt{description}:]
{\bf An account of the content of the resource.}
Description may include but is not limited to: an abstract, table of
contents, reference to a graphical representation of content, or a free-text
account of the content.

\item[\elt{format}:]
{\bf The physical or digital manifestation of the resource.}
Typically, \elt{format} will specify the media-type or dimensions of a
physical resource, or the character encoding or markup of a
digital resource.  It may be used to determine the software, hardware or other
equipment needed to use the resource.
Since this element applies both to software and data, service
providers can use it to match data with appropriate software tools
and vice versa.

\item[\elt{identifier}:]
{\bf An unambiguous reference to the resource within a given context.}
Recommended best practice is to identify the resource by means of a
string or number conforming to a globally-known formal identification system
(e.g.~by URI or ISBN).
For non-digital archives, \elt{identifier} may use
the existing scheme for locating a resource within the collection.

\item[\elt{language}:]
{\bf A language of the intellectual content of the resource.}
The \elt{language} element is used for a language the resource is in,
as opposed to a language it describes (i.e.~a ``subject language'').
It identifies a language that the creator of the
resource assumes that its eventual user will understand.
Recommended best practice is to identify the language precisely
using a coded value from the OLAC Language Vocabulary.

\item[\elt{publisher}:]
{\bf An entity responsible for making the resource available.}
Examples of a publisher include a person, an organization, or a
service.

\item[\elt{relation}:]
{\bf A reference to a related resource.}
This element is used to document relationships between resources.
Dublin Core qualifiers may be used to refine the nature
of the relationship (for instance, is
replaced by, requires, is part of, and so on).
The refinements to \elt{relation} are defined in \citep{DCER02}.

\item[\elt{rights}:]
{\bf Information about rights held in and over the resource.}
Typically, a \elt{rights} element will contain a rights management
statement for the resource, or reference a service providing such information.
Rights information often encompasses intellectual property rights,
copyright, and various property rights.

\item[\elt{source}:]
{\bf A reference to a resource from which the present resource is derived.}
For instance, it
may be the bibliographic information about a printed book of which this is the
electronic encoding or from which the information was extracted.

\item[\elt{subject}:]
{\bf The topic of the content of the resource.}
Typically, a Subject will be expressed as keywords, key phrases or
classification codes that describe a topic of the resource. Recommended best
practice is to select a value from a controlled vocabulary or formal
classification scheme.  Where the subject of the resource is a language,
recommended best practice is to use the
OLAC Language Vocabulary (cf.~the \elt{language} element above).

\item[\elt{title}:]
{\bf A name given to the resource.}
Typically, a title will be a name by which the resource is formally known.

\item[\elt{type}:]
{\bf The nature or genre of the content of the resource.}
Recommended best practice is to use the
Dublin Core controlled vocabulary DC-Type for broad classification
of type.  OLAC provides additional vocabularies that are relevant for
language resources:
the OLAC Linguistic Data Type Vocabulary \citep{OLAC-Type}, and
the OLAC Discourse Type Vocabulary \citep{OLAC-Discourse}.

\end{description}

\subsection{The controlled vocabularies}
\label{sec:cv}

Controlled vocabularies are enumerations of legal values, or specifications
of legal formats, for the \attr{code} attribute.
In some cases, more than one value applies,
in which case the corresponding element must be repeated, once for each
applicable value.  In other cases, no value is applicable ands
the corresponding element is simply omitted.  In yet other cases, the
controlled vocabulary may fail to provide a suitable item, in which case
a similar item can be optionally specified and a prose comment included in the
element content.

\subsubsection{The OLAC Language Vocabulary}

Language identification is an important dimension of language resource
classification. However, the character-string representation of language names
is problematic for several reasons:
different languages (in different parts of the world) may have the
same name;
the same language may have a different name in each country where
it is spoken;
within the same country, the preferred name for a language may
change over time;
in the early history of discovering new languages (before names
were standardized), different people referred to the same language by different
names; and
for languages having non-Roman orthographies, the language name
may have several possible romanizations.
Together, these facts suggest that a standard based
on names will not work.
Instead, we need a standard based on unique identifiers
that do not change, combined with accessible documentation that
clarifies the particular speech variety denoted by each identifier.

The information technology community has a standard for language
identification, namely, ISO 639 \citep{ISO639}. Part 1 of this standard
lists two-letter codes for identifying 160 of the world's major
languages; part 2 of the standard lists three-letter codes for identifying
about 400 languages. ISO 639 in turn forms the core of another standard, RFC
3066 (formerly RFC 1766), which is the
standard used for language identification in the xml:lang attribute of XML and
in the language element of the Dublin Core metadata set.  RFC 3066
provides a mechanism for users to register new language identification codes
for languages not covered by ISO 639, but very few additional languages have
ever been registered.

Unfortunately, the existing standard falls far short of meeting the
needs of the language resources community since it fails to account for more
than 90\% of the world's languages, and it fails to adequately document what
languages the codes refer to \citep{Simons00}. However, SIL's Ethnologue
\citep{Grimes00} provides a complete system of language identifiers which
is openly available on the Web. OLAC will employ the RFC 3066 extension
mechanism to build additional language identifiers based on the Ethnologue
codes.  For the 130-plus ISO-639-1 codes having a one-to-one mapping onto
Ethnologue codes, OLAC will support both.  Where an ISO code is ambiguous
OLAC requires the Ethnologue code.
New identifiers for ancient languages, currently being developed by
LINGUIST List, will be incorporated.
These language identifiers are expressed using the \attr{code} attribute of the
\elt{language} and \elt{subject} elements (using the special
\attr{x-} prefix of RFC 3066 for user-defined extensions).
The free-text content of these elements may be used to specify an
alternative human-readable name for the language (where the name
specified by the standard is unacceptable for some reason)
or to specify a dialect (where the resource is dialect-specific).

\subsubsection{The OLAC Linguistic Data Type Vocabulary}

After language identification, another dimension of central importance
for language resources is 
the linguistic type of a resource.  Notions such as ``lexicon''
and ``primary text'' are fundamental, and the discourse of the
language resources community depends on shared assumptions about what
these types mean.

At present, the OLAC Linguistic Data Type Vocabulary \citep{OLAC-Type}
distinguishes just three types: \code{lexicon}, \code{primary\_text},
and \code{language\_description}.  A lexicon is defined as a
``systematic listing of lexical entries...  Each lexical item may, but
need not, be accompanied by a definition, a description of the
referent (in the case of proper names), or an indication of the item's
semantic relationship to other lexical items.''  A primary text is
defined as ``linguistic material which is itself the object of study,
typically material in the subject language which is a performance of a
speech event, or the written analog of such an event.''
Finally, language description is a
resource which ``describes a language or some aspect(s) of a language
via a systematic documentation of linguistic structures.''

\subsubsection{Other controlled vocabularies}

Here we list three other OLAC vocabularies.  For full definitions,
examples and notes, the reader is referred to the cited vocabulary
document.

\begin{description}\setlength{\itemsep}{0pt}\setlength{\parskip}{0pt}
\item[Discourse Type:]
The OLAC Discourse Type Vocabulary
describes ``the content of a resource as representing discourse
of a particular structural type''
\citep{OLAC-Discourse}.  The vocabulary terms are as follows:
drama, formulaic discourse, interactive discourse, language play,
oratory, narrative, procedural discourse, report, singing, and
unintelligible speech.

\item[Role:]
The OLAC Role Vocabulary
\citep{OLAC-Role} serves to identify the role of an individual
or institution in creating or contributing to a language resource.
The vocabulary terms are as follows:
annotator, artist, author, compiler, consultant, depositor, developer,
editor, illustrator, interviewer, participant, performer,
photographer, recorder, researcher, respondent, signer, speaker,
sponsor, transcriber, and translator.

\item[Linguistic Subject:] The OLAC Linguistic Subject Vocabulary
\citep{OLAC-Subject} describes the content of a resource as being
about a particular subfield of linguistic science.  The list has been
developed in the course of classifying resources on the LINGUIST List
website.  The vocabulary terms are as follows:
anthropological linguistics, applied linguistics, cognitive science,
computational linguistics, discourse analysis, forensic linguistics,
general linguistics, historical linguistics, history of linguistics,
language acquisition, language documentation, lexicography,
linguistics and literature, linguistic theories, mathematical
linguistics, morphology, neurolinguistics, philosophy of language,
phonetics, phonology, pragmatics, psycholinguistics, semantics,
sociolinguistics, syntax, text and corpus linguistics, translating and
interpreting, typology, and writing systems.
\end{description}

In addition to the five vocabularies discussed here, other
vocabularies have been proposed and are being considered by
the community.

Once a vocabulary is reviewed and accepted by the community as OLAC
best practice in language resource description, the corresponding XML
schema is hosted on the OLAC website.  Archives which use this
vocabulary can then be automatically tested for conformance.
Prior to acceptance, any new vocabulary can be set up as a
``third-party extension'' and adopted by archives without any
centralized review process.  This bottom-up approach encourages
experimentation and innovation, yet only leads to community-wide
adoption once the benefit of the new vocabulary for resource discovery
has been demonstrated.

\subsection{XML representation}
\label{sec:xml}

The XML implementation of OLAC metadata follows the ``Guidelines for
implementing Dublin Core in XML'' \citep{DCXML03}. The OLAC metadata schema is
an application profile \citep{HeeryPatel00} that incorporates the elements from
two metadata schemas developed by the DC Architecture Working Group
for implementing qualified DC.
The most recent version of the OLAC metadata schema
is posted on the OLAC website\footnote{\url{http://www.language-archives.org/OLAC/1.0/olac.xsd}}
and an example record is available\footnote{\url{http://www.language-archives.org/OLAC/1.0/olac.xml}}.

The container for an OLAC metadata record is the element
\elt{olac}, which
is defined in a namespace called
\url{http://www.language-archives.org/OLAC/1.0/}. By convention the
namespace prefix \elt{olac} is used, and the DC namespace is declared to be
the default so that the metadata element tags need not be prefixed.
For instance, the following is a valid OLAC metadata record:

\begin{alltt}\small
<olac:olac
   xmlns:olac="http://www.language-archives.org/OLAC/1.0/"
   xmlns="http://purl.org/dc/elements/1.1/"
   xmlns:xsi="http://www.w3.org/2001/XMLSchema-instance"
   xsi:schemaLocation=
      "http://www.language-archives.org/OLAC/1.0/ 
       http://www.language-archives.org/OLAC/1.0/olac.xsd">
   <creator>Bloomfield, Leonard</creator>
   <date>1933</date>
   <title>Language</title>
   <publisher>New York: Holt</publisher>
</olac:olac>
\end{alltt}

In addition to this DC metadata, an element may use a DC qualifier,
following the guidelines given in \citep{DCXML03}. The element may specify a
refinement (using an element defined in the dcterms namespace) or an
encoding scheme (using a scheme defined in \elt{dcterms} as the value of the
\attr{xsi:type} attribute), or both. Note that the metadata record must
declare the \elt{dcterms} namespace as follows:
\attr{xmlns:dcterms="http://purl.org/dc/terms/"}. For instance, the following
element represents a creation date encoded in the W3C date and time
format:

\begin{alltt}\small
<dcterms:created xsi:type="dcterms:W3C-DTF">2002-11-28
  </dcterms:created>
\end{alltt}

The \attr{xsi:type} attribute is a directive that is built into the
XML Schema standard [\url{http://www.w3.org/XML/Schema}].
It functions to override the type definition of
the current element by the type definition named in its value. In this
example, the value of \attr{dcterms:W3C-DTF} resolves to a complex type
definition in the XML schema for the \attr{dcterms} namespace.

Any element may also use the \attr{xml:lang} attribute to indicate the
language of the element content. For instance, the following
represents a title in the Lau language of Solomon Islands and its
translation into English:

\begin{alltt}\small
<title xml:lang="x-sil-LLU">Na tala 'uria na idulaa
  diana</title>
<dcterms:alternative xml:lang="en">The road to good
  reading</dcterms:alternative>
\end{alltt}

For further detailed discussion of the XML format,
the reader is referred to
\citep{SimonsBird03lht,SimonsBird03metadata}.

\subsection{Mapping OLAC metadata to other formats}

As we have seen, OLAC metadata uses attributes to support resource
description using controlled vocabularies, and service providers may
use these attributes to perform precise searches. However, service
providers also need to be able to display metadata records to users in
an easy-to-read format.  This involves translating coded attribute
values into human-readable form, and combining this information with
the element content to produce a display of all information pertaining
to a metadata element \citep{Simons03display}.

Transforming OLAC metadata records into such a display format is a
non-trivial task.  Instead of having each service provider perform this
task independently, OLACA, the OLAC Aggregator \citep{SimonsBird03lht}
offers a human-readable version of all OLAC metadata.
Service providers can harvest this metadata, and
expose the content of the metadata elements to end-users without any further
processing.

Beyond this, the OLAC website exposes human-readable versions of OLAC
metadata to wider communities.  First, a simple DC version
of the human-readable metadata is exposed to OAI service providers, so
that all OLAC archives show up in digital library catalogs of the
wider OAI community (e.g. in the ARC
service \url{http://arc.cs.odu.edu/}).  Second, an HTML version of the
human-readable metadata is exposed to web crawlers, permitting all
OLAC metadata records to be indexed by web search engines and to be stored in
internet archives.

\section{Conclusions}

As language resources proliferate, and as the associated community
grows, the need for a consistent and comprehensive framework for
resource description and discovery is becoming critical.  OLAC has
addressed this need by providing metadata tailored to the needs of
language resource description, minimally extending the DC standard.
At the same time, the OAI Protocol for Metadata Harvesting on which the
OLAC infrastructure is built permits
end-users to search the contents of multiple archives from a single location.

OLAC provides a ready \emph{template} for resource
description, with two clear benefits over traditional full-text description
and retrieval.  First, the template guides the resource creator in giving a
\emph{complete description} of the resource, in contrast to prose
descriptions which may omit important details.  And second, the template
associates the elements of a description
with \emph{standard labels}, such as \elt{creator}
and \elt{title}, permitting users to do focussed searching.  Resources and
repositories can proliferate, yet a common metadata format
will support centralized services, giving users easy access to language resources.

Despite its many benefits, simply making resources findable is
insufficient on its own.  There must also be a framework in which the
community can identify and promote best practices for digital representation
of linguistic information to ensure re-usability and long-term
preservation.  To support this need, OLAC has developed a process
which specifies how the community can identify best practices
\citep{OLAC-Process}.

We conclude by calling for wider participation in OLAC.  First, the
controlled vocabularies used by the OLAC Metadata Set and described in this
article are works in progress, and are continuing to be revised with input
from participating archives and members of the community.  We hope to have
provided sufficient motivation and exemplification for
readers to be able to contribute to ongoing developments.  Second,
the OLAC process can be used by community members to develop new
vocabularies and other best practice recommendations.
Finally, the core infrastructure of data
providers and service providers is operational, and individuals and
institutions are encouraged to use it for the widespread dissemination
of their language resources.

\section*{Acknowledgements}

This material is based upon work supported by the National Science
Foundation under grants:
9910603 \emph{International Standards in Language Engineering},
and 9978056 \emph{TalkBank}.
Earlier versions of this material were presented at the
Workshop on Web-Based Language Documentation and Description
in Philadelphia, December 2000 \citep{BirdSimons00},
the ACL/EACL Workshop on Sharing Tools and Resources for
Research and Education \citep{BirdSimons01}, and the
IRCS Workshop on Open Language Archives \citep{BirdSimons02workshop}.
We are indebted to
members of the OLAC community for their active participation in
the creation and development of the OLAC metadata format.

\def\UrlFont{\small\tt}
\raggedright


\begin{thebibliography}{23}
\expandafter\ifx\csname natexlab\endcsname\relax\def\natexlab#1{#1}\fi
\expandafter\ifx\csname url\endcsname\relax
  \def\url#1{{\tt #1}}\fi

\bibitem[{Aristar Dry} and Appleby(2003)]{OLAC-Subject}
Helen {Aristar Dry} and Michael Appleby.
\newblock {OLAC} linguistic subject vocabulary, 2003.
\newblock \url{http://www.language-archives.org/REC/field.html}.

\bibitem[{Aristar Dry} and Johnson(2002)]{OLAC-Type}
Helen {Aristar Dry} and Heidi Johnson.
\newblock {OLAC} linguistic data type vocabulary, 2002.
\newblock \url{http://www.language-archives.org/REC/type.html}.

\bibitem[Bird and Simons(2000)]{BirdSimons00}
Steven Bird and Gary Simons, editors.
\newblock {\em Proceedings of the Workshop on Web-Based Language Documentation
  and Description}, 2000.
\newblock \url{http://www.ldc.upenn.edu/exploration/expl2000/}.

\bibitem[Bird and Simons(2001)]{BirdSimons01}
Steven Bird and Gary Simons.
\newblock The {OLAC} metadata set and controlled vocabularies.
\newblock In {\em Proceedings of ACL/EACL Workshop on Sharing Tools and
  Resources for Research and Education}, 2001.
\newblock \url{http://arXiv.org/abs/cs/0105030}.

\bibitem[Bird and Simons(2002)]{BirdSimons02workshop}
Steven Bird and Gary Simons, editors.
\newblock {\em Proceedings of the IRCS Workshop on Open Language Archives},
  2002.
\newblock \url{http://www.language-archives.org/events/olac02/}.

\bibitem[Bird and Simons(2003)]{BirdSimons03language}
Steven Bird and Gary Simons.
\newblock Seven dimensions of portability for language documentation and
  description.
\newblock {\em Language}, 79:\penalty0 557--82, 2003.

\bibitem[{DCMI}(2000)]{DCQ00}
{DCMI}.
\newblock {Dublin Core} qualifiers, 2000.
\newblock \url{http://dublincore.org/documents/2000/07/11/dcmes-qualifiers/}.

\bibitem[{DCMI}(2002)]{DCER02}
{DCMI}.
\newblock {DCMI} elements and element refinements -- a current list, 2002.
\newblock \url{http://dublincore.org/usage/terms/dc/current-elements/}.

\bibitem[Grimes(2000)]{Grimes00}
Barbara~F. Grimes, editor.
\newblock {\em Ethnologue: Languages of the World}.
\newblock Dallas: Summer Institute of Linguistics, 14th edition, 2000.
\newblock \url{http//www.ethnologue.com/}.

\bibitem[Heery and Patel(2000)]{HeeryPatel00}
Rachel Heery and Manjula Patel.
\newblock Application profiles: mixing and matching metadata schemas.
\newblock In {\em Ariadne}, volume~25. UK Office for Library and Information
  networking (UKOLN), University of Bath, 2000.
\newblock \url{http://www.ariadne.ac.uk/issue25/app-profiles/}.

\bibitem[{ISO}(1998)]{ISO639}
{ISO}.
\newblock {ISO} 639: Codes for the representation of names of languages-part 2:
  Alpha-3 code, 1998.
\newblock \url{http://lcweb.loc.gov/standards/iso639-2/langhome.html}.

\bibitem[Johnson(2002)]{OLAC-Role}
Heidi Johnson.
\newblock {OLAC} role vocabulary, 2002.
\newblock \url{http://www.language-archives.org/REC/role.html}.

\bibitem[Johnson and {Aristar Dry}(2002)]{OLAC-Discourse}
Heidi Johnson and Helen {Aristar Dry}.
\newblock {OLAC} discourse type vocabulary, 2002.
\newblock \url{http://www.language-archives.org/REC/discourse.html}.

\bibitem[Lagoze and {Van de Sompel}(2001)]{LagozeVandeSompel01}
Carl Lagoze and Herbert {Van de Sompel}.
\newblock The {Open Archives Initiative}: Building a low-barrier
  interoperability framework.
\newblock In {\em Proceedings of the First ACM/IEEE-CS Joint Conference on
  Digital Libraries}, pages 54--62, 2001.
\newblock \url{http://www.cs.cornell.edu/lagoze/papers/oai-jcdl.pdf}.

\bibitem[Powell and Johnston(2003)]{DCXML03}
Andy Powell and Pete Johnston.
\newblock Guidelines for implementing {Dublin Core} in {XML}, 2003.
\newblock \url{http://dublincore.org/documents/dc-xml-guidelines/}.

\bibitem[Simons(2000)]{Simons00}
Gary Simons.
\newblock Language identification in metadata descriptions of language archive
  holdings.
\newblock In Steven Bird and Gary Simons, editors, {\em Proceedings of the
  Workshop on Web-Based Language Documentation and Description}, 2000.
\newblock \url{http://www.ldc.upenn.edu/exploration/expl2000/papers/simons/}.

\bibitem[Simons(2003)]{Simons03display}
Gary Simons.
\newblock Specifications for an olac metadata display format and an
  olac-to-oai\_dc crosswalk, 2003.
\newblock \url{http://www.language-archives.org/NOTE/olac_display.html}.

\bibitem[Simons and Bird(2002)]{OLAC-Process}
Gary Simons and Steven Bird.
\newblock {OLAC} process, 2002.
\newblock \url{http://www.language-archives.org/OLAC/process.html}.

\bibitem[Simons and Bird(2003{\natexlab{a}})]{SimonsBird03lht}
Gary Simons and Steven Bird.
\newblock Building an {Open Language Archives Community} on the {OAI}
  foundation.
\newblock {\em Library Hi Tech}, 21:\penalty0 210--218, 2003{\natexlab{a}}.
\newblock \url{http://www.arxiv.org/abs/cs.CL/0302021}.

\bibitem[Simons and Bird(2003{\natexlab{b}})]{SimonsBird03metadata}
Gary Simons and Steven Bird.
\newblock {OLAC} metadata, 2003{\natexlab{b}}.
\newblock \url{http://www.language-archives.org/OLAC/metadata.html}.

\bibitem[Svenonius(2000)]{Svenonius00}
Elaine Svenonius.
\newblock {\em The Intellectual Foundation of Information Organization}.
\newblock The MIT Press, 2000.

\bibitem[{Van de Sompel} and Lagoze(2002)]{VandeSompelLagoze02}
Herbert {Van de Sompel} and Carl Lagoze.
\newblock Notes from the interoperability front: A progress report on the {Open
  Archives Initiative}.
\newblock In {\em Proceedings of the European Conference on Digital Libraries},
  pages 144--157, 2002.
\newblock \url{http://www.openarchives.org/documents/ecdl-oai.pdf}.

\bibitem[Wolf and Wicksteed(1997)]{W3CDTF}
Misha Wolf and Charles Wicksteed.
\newblock Date and time formats, 1997.
\newblock \url{http://www.w3.org/TR/NOTE-datetime}.

\end{thebibliography}
\end{document}